\documentclass[10pt,twocolumn,letterpaper]{article}

\usepackage[pagenumbers]{cvpr} 

\usepackage{graphicx}
\usepackage{amsmath}
\usepackage{amssymb}
\usepackage{booktabs}
\usepackage{multirow}
\usepackage{pifont}
\usepackage{xspace}
\usepackage{array}
\usepackage{makecell}
\usepackage{tabularx}
\usepackage{adjustbox}
\newcommand{\cmark}{\ding{51}}

\usepackage[accsupp]{axessibility}  

\usepackage[pagebackref,breaklinks,colorlinks]{hyperref}

\usepackage[capitalize]{cleveref}
\crefname{section}{Sec.}{Secs.}
\Crefname{section}{Section}{Sections}
\Crefname{table}{Table}{Tables}
\crefname{table}{Tab.}{Tabs.}

\def\confName{CVPR}
\def\confYear{2022}
\begin{document}
\title{Threshold Matters in WSSS: Manipulating the Activation for the Robust and Accurate Segmentation Model Against Thresholds}

\author{Minhyun Lee\thanks{indicates an equal contribution.}, \ Dongseob Kim$^*$, \ Hyunjung Shim\thanks{Hyunjung Shim is a corresponding author.}\\
Yonsei University\\
{\tt\small \{lmh315, kou.k, kateshim\}@yonsei.ac.kr}
}
\maketitle
\begin{abstract}
Weakly-supervised semantic segmentation (WSSS) has recently gained much attention for its promise to train segmentation models only with image-level labels. Existing WSSS methods commonly argue that the sparse coverage of CAM incurs the performance bottleneck of WSSS. This paper provides analytical and empirical evidence that the actual bottleneck may not be sparse coverage but a global thresholding scheme applied after CAM. Then, we show that this issue can be mitigated by satisfying two conditions; 1) reducing the imbalance in the foreground activation and 2) increasing the gap between the foreground and the background activation. Based on these findings, we propose a novel activation manipulation network with a per-pixel classification loss and a label conditioning module. Per-pixel classification naturally induces two-level activation in activation maps, which can penalize the most discriminative parts, promote the less discriminative parts, and deactivate the background regions. Label conditioning imposes that the output label of pseudo-masks should be any of true image-level labels; it penalizes the wrong activation assigned to non-target classes. Based on extensive analysis and evaluations, we demonstrate that each component helps produce accurate pseudo-masks, achieving the robustness against the choice of the global threshold. Finally, our model achieves state-of-the-art records on both PASCAL VOC 2012 and MS COCO 2014 datasets. The code is available at \href{https://github.com/gaviotas/AMN}{https://github.com/gaviotas/AMN}.
\end{abstract}
\section{Introduction}
\label{sec:intro}

Weakly-supervised semantic segmentation (WSSS) requires only weak supervision (e.g., image-level labels~\cite{pathak2015constrained,pinheiro2015image}, scribbles~\cite{lin2016scribblesup}, bounding boxes~\cite{khoreva2017simple}) as opposed to the fully supervised model, which involves costly pixel-level annotations. In this work, we address WSSS using image-level labels because of its low labeling cost. The overall pipeline of WSSS consists of two stages. The pseudo-mask is first generated from an image classifier, and then it is used as supervision to train a segmentation network. 

The prevalent technique for generating pseudo-masks is class activation mapping (CAM)~\cite{zhou2016learning}. It uses the intermediate classifier's activation to compute the class activation map corresponding to its image-level label. The common practice of WSSS is to apply a global threshold to the activation map (i.e., assigning the object class if the activation is greater than the threshold) for obtaining the pseudo-mask. Existing methods point out that the pseudo-mask obtained from CAM only captures the most discriminative parts of the object, incurring the performance bottleneck. Therefore, most existing studies have expanded object coverage by manipulating the image~\cite{li2018tell,wei2017object} or feature map~\cite{lee2019ficklenet,jiang2019integral}.

\begin{figure}[t]
\centering
\includegraphics{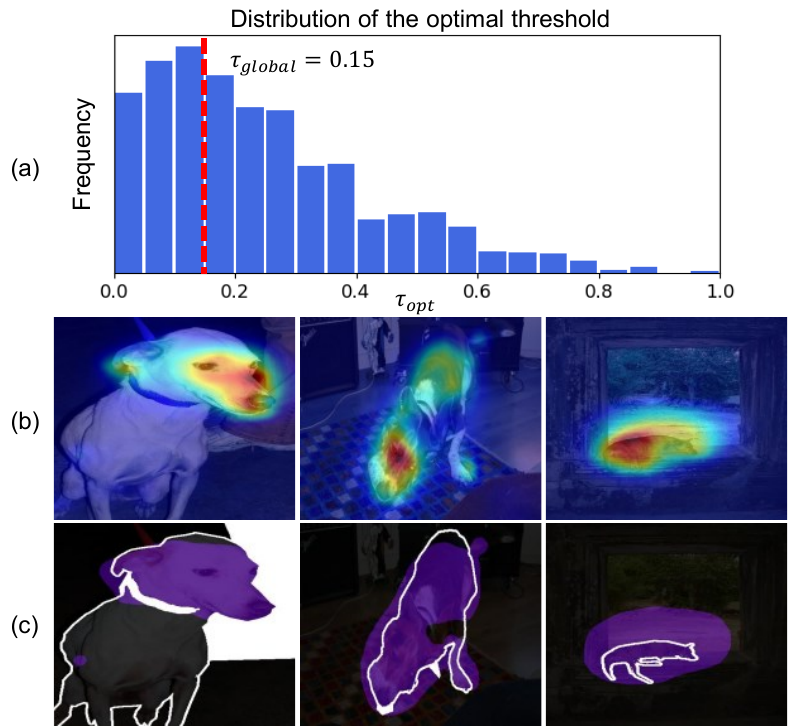}
\vspace{-1mm}
\caption{Motivating examples show that the optimal threshold per image ($\tau_{opt}$) from the same \emph{dog} class is quite different from each other. (a) The distribution of the optimal threshold on PASCAL VOC 2012 train set, (b) the activation maps, (c) the thresholded masks using a global threshold $\tau_{global}=0.15$.}
\label{fig:concept}
\vspace{-5mm}
\end{figure}

However, we argue that the performance bottleneck of WSSS comes from a global threshold applied after CAM; the sparse object coverage does not explain all. This threshold partitions each activation map into the foreground (object class) and background. Then, the pseudo-mask is generated by combining all foreground regions. Here, the choice of threshold critically affects the performance of WSSS. i) We further observe that a global threshold cannot provide an optimal threshold per image. Figure~\ref{fig:concept}(a) visualizes the distribution of optimal threshold on the PASCAL VOC 2012 train set (For analysis, we obtain the best threshold per image using its ground-truth segmentation map). It shows that the optimal threshold per image quite differs from each other, and the global threshold is often far from the optimal one. ii) Besides, a global threshold for CAM does not always lead to sparse coverage. Figure~\ref{fig:concept}(b) and (c) show several CAM examples and the corresponding masks generated by a global threshold, respectively; the third row shows that CAM and the thresholded mask overly capture the target object. These results clearly motivate us to rethink that the performance bottleneck of WSSS is closely related to the usage of a global threshold.

To tackle this problem, we first investigate why this problem happens. By tracing the procedure of CAM, we realize that global average pooling (GAP) applied to the last layer invokes this issue; the global threshold largely differs from the optimal threshold per image. The first stage of the WSSS framework trains the image classifier, whose score is computed via GAP. While GAP facilitates deriving the activation map, it averages the feature maps into a single classification score. The same value can be from totally different activations. For example, the same score can be from 1) high activations only in the most discriminative region (low optimal threshold to cover more regions), 2) moderate activations distributed over the entire object, or 3) small activations covering even outside the object (high optimal threshold to cover small regions).

Due to its averaging nature, GAP hinders achieving the accurate pseudo-mask via a global threshold. As a na\"ive solution, one might consider introducing a different threshold per image. However, this is prohibitive because finding a per-image threshold requires pixel-level annotation, violating the principle of weakly-supervised learning. Instead of controlling a threshold per image, our key idea is to manipulate the activation in a way that the resultant pseudo-mask is of high quality regardless of threshold values. To achieve robust performance, we can increase the activation gap between the foreground region and the background region; the thresholded masks are the same if the threshold value is within the gap. However, it can induce the model to capture the most discriminative parts only, resulting in consistent but poor quality. 

To achieve high quality consistently, it is important to reduce the activation imbalance within the foreground and keep the large activation gap between the foreground and background simultaneously. We can achieve the two factors by assigning the two-level activation for the entire foreground pixels and background pixels (e.g., 1 and 0). In this way, the high activation in the most discriminative parts is penalized, but the low activation in the less discriminative parts is promoted. Meanwhile, the background activation can be deactivated. Naturally, this strategy can guarantee a large gap, enabling us to achieve the robust performance even with a global threshold.

Specifically, we introduce a robust and accurate \emph{activation manipulation network (AMN)}, which takes an image with its image-level label as the input and provides the high-quality pseudo-mask as the output. For that, we formulate a training objective using i) per-pixel classification with an effective constraint using ii) label conditioning. Since per-pixel classification does not rely on GAP, it bypasses the issue of having totally different activation maps for the same classification score. More importantly, it directly enforces the same large activation for the foreground (e.g., 1) and the same small activation for the background (e.g., 0). Here, it leads to reducing the activation imbalance inside the foreground and having a large gap between the foreground and background. Since we do not have pixel-level supervision to formulate per-pixel classification problems, the noisy pseudo-mask from CAM with conditional random field (CRF)~\cite{krahenbuhl2011efficient} serves as the initial target for training AMN. 

Moreover, we propose label conditioning to reduce the activation of non-target classes. The idea of label conditioning is to reformulate the label prediction problem by finding the best prediction out of the given $K$ classes ($K$ is the number of classes given by the ground-truth image-level label per image) and background, instead of $N+1$ classes (i.e., a total of $N$ foreground classes and a background class). $K$ is always much less than $N$ and thereby the range of possible answers is clearly reduced. It makes the problem better-posed. More importantly, the activation of non-target classes is strictly suppressed by mapping to 0. It helps strengthen the foreground activation. As a result, with the same global threshold as the previous studies~\cite{ahn2019weakly, lee2021anti, lee2021reducing}, AMN largely improves the quality of pseudo-mask and eventually records the state-of-the-art performances on the Pascal VOC 2012 and MS COCO 2014 benchmarks.

\section{Related work}
\label{sec:related}

Most WSSS techniques utilize CAM~\cite{zhou2016learning} to obtain localization maps from image-level labels. Considering the sparse coverage of CAM as the bottleneck of WSSS, many studies focused on expanding the seed activation of CAM. Specifically, \cite{li2018tell,hou2018self,choe2020attention} suggested erasing the most discriminative regions. CIAN~\cite{fan2020cian} utilized the cross-image affinity. \cite{chang2020weakly,wang2020self} devised a self-supervised task. \cite{lee2019ficklenet, jiang2019integral} suggested a feature ensemble method. \cite{LiZLZZ21Group,sun2020mining} developed a class-wise co-attention mechanism. AdvCAM~\cite{lee2021anti} proposed an anti-adversarial image manipulation method. \cite{ahn2018learning,ahn2019weakly} implicitly exploit the boundary information with pixel-level affinity information, naturally expanding the object coverage until boundaries. 

Another approach exploits additional information to refine the object boundaries or distinguish co-occurring objects~\cite{wei2016stc,fan2020employing,zeng2019joint}. \cite{chaudhry_dcsp_2017,yao2020saliency} combined saliency maps with class-specific attention cues to generate reliable pseudo-masks. EPS~\cite{Lee_2021_CVPR} utilized the saliency maps as the cues for boundaries and co-occurring pixels. DRS~\cite{KimHK21DRS} suppressed the most discriminative parts to expand the object coverage and then refine the boundary with saliency map.

Several existing studies resolved the limitation of CAM-GAP by modifying the pooling methods~\cite{kolesnikov2016seed, araslanov2020single, lee2021reducing}. SEC~\cite{kolesnikov2016seed} argued that global max pooling (GMP) underestimates the object size and GAP sometimes overestimates it. Then, they proposed global weighted rank pooling. Araslanov et al.~\cite{araslanov2020single} claimed that CAM-GAP may penalize small segments and proposed normalized global weighted pooling (nGWP) instead of GAP. As a concurrent work, RIB~\cite{lee2021reducing} promotes less discriminative regions by collecting only non-discriminative regions for pooling. Unlike the methods suggested new pooling layers, we focus on the fact that GAP (in fact, any pooling methods will do) leads to having a different optimal threshold per image.
\section{Preliminaries}\label{section:3}

\noindent\textbf{Class activation mapping (CAM).} In the WSSS framework, CAM is used to provide class activation maps corresponding to their image-level labels. Given a CNN $f$ and the input $x \in \mathbb{R}^{H \times W \times 3}$, $H$ and $W$ indicate the height and width of the input. The feature maps are average pooled and then multiplied by the weights $\mathbf{w}_i^{c}$ for class $c$ from the classifier, resulting in the classification score. By multiplying $\mathbf{w}_i^c$ back to the feature maps $f(x) \in \mathbb{R}^{H_{out} \times W_{out} \times Q}$, we can compute the class activation map $\mathrm{F}_c(x) \in \mathbb{R}^{H_{out} \times W_{out}}$ for class $c$ as follows:
\vspace{-2mm}
\begin{equation}
{\small
    \mathrm{F}_c(x) = \sum_{i=1}^{Q}\mathbf{w}_i^{c \top} \cdot f_i(x),    
}
\end{equation}\vspace{-2mm}

\noindent where $Q$ is the number of channels in feature maps. 

All existing WSSS methods normalize $\mathrm{F}_c$ into the range of [0 1] and then apply a global threshold to separate the foreground and background pixels. In this way, we can generate pixel-level masks from image-level labels.

\begin{figure}[t]
\centering
\includegraphics{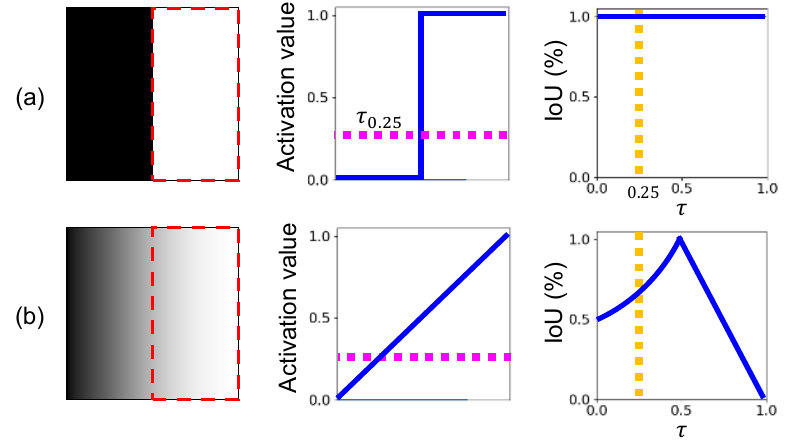}
\vspace{-2mm}
\caption{Two different activation maps and their IoU curves under different thresholds. The black with 0 activation for the background, the white with 1 activation for the foreground, and a red box for the ground-truth object region. A magenta and yellow indicate a threshold line at $\tau=0.25$ and corresponding IoU of each activation map, respectively. The plots in the second column are the one-dimensional horizontal cross-section of the activation map.}


\label{fig:th_robust}
\vspace{-5mm}
\end{figure}

\vspace{1mm}
\noindent\textbf{Threshold matters in WSSS.} GAP allows different activations to be mapped to the same classification score. Thus, the resultant $\mathrm{F}_c$ can have various distributions of activations. As a result, as seen in Figure~\ref{fig:concept}, no single threshold can be sufficient to derive the optimal pseudo-masks for different inputs. We investigate the conditions where the pseudo-mask is accurate and robust against different thresholds. The first condition (\textcolor{blue}{c1}) is reducing the activation imbalance within the foreground as also pointed out in~\cite{Kweon_2021unlocking}. It guides the activation value cover the entire extent of the target object rather than focusing on the most discriminative part. The second condition (\textcolor{blue}{c2}) is enforcing the large activation gap between the foreground and the background activation. It helps the pseudo-mask generation less sensitive against the threshold. By jointly satisfying \textcolor{blue}{c1} and \textcolor{blue}{c2}, we argue that the activation is formed to distinguish the foreground and the background reasonably well with a global threshold. A simple toy example in Figure~\ref{fig:th_robust}(a) illustrates that satisfying the two conditions can guarantee consistent and accurate performance regardless of threshold $\tau$; the same pseudo-mask is generated by choosing any $\tau$ within the gap. Figure~\ref{fig:th_robust}(b) shows the opposite scenario, where it satisfies neither \textcolor{blue}{c1} nor \textcolor{blue}{c2}; the performance is extremely sensitive to the choice of $\tau$. The two cases illustrate that satisfying both \textcolor{blue}{c1} and \textcolor{blue}{c2} allows us to obtain accurate and robust pseudo-masks.

\section{Activation Manipulation Network}

\begin{figure*}
\begin{center}
\includegraphics[]{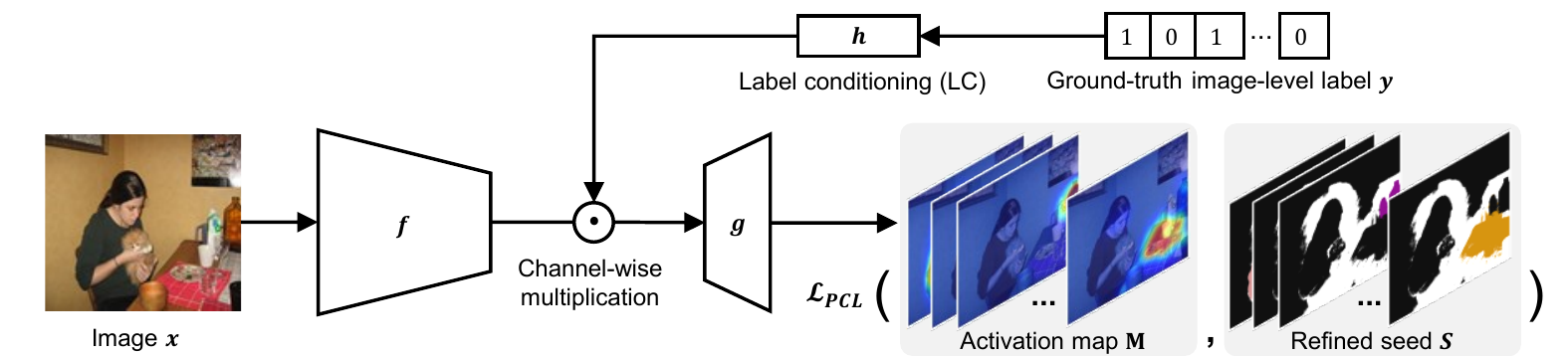}
\end{center}
\vspace{-3mm}
\caption{The overall framework of activation manipulation network (AMN). The refined seed from a classification network is used as noisy supervision to train AMN. The per-pixel classification loss (PCL) and label conditioning (LC) improve the pseudo-mask quality.}
\label{fig:overview}
\vspace{-4mm}
\end{figure*}
Our goal is to improve pseudo-mask quality by manipulating the activation map at the pixel-level, leading to robust performance against threshold choice. To this end, we propose an \textit{activation manipulation network} (AMN) with two learning objectives. Firstly, we introduce a \textit{per-pixel classification loss}, which reduces the activation imbalance inside foreground and provides the large gap between the foreground and the background (i.e., 1 when it is normalized into [0 1]). In addition, we propose a \textit{label conditioning} module, which eliminates the activation from the non-target classes. It helps produce the foreground and background activation more accurately by reallocating the activation.

\subsection{Overall training procedure}\label{subsection:4.1}

The training of the proposed WSSS framework consists of three stages: 1) seed generation, 2) pseudo-mask generation with the proposed AMN, and 3) final segmentation. For seed generation, we obtain noisy pixel-level annotations from image-level labels by applying CAM. Then, we apply conditional random field (CRF)~\cite{krahenbuhl2011efficient}. CRF is the prevalent post-processing method and refines the initial seed by assigning an undefined region for less confident pixels. Specifically, we follow the procedure of Ahn et al.~\cite{ahn2019weakly} to generate the refined seed $\mathbf{S}_{i}$. Secondly, given the image $\mathbf{x}_{i}$ and its image-level label $\textbf{y}_{i}$ as inputs and the refined seed $\mathbf{S}_{i}$ as the target output, we train AMN via a per-pixel classification loss (PCL) with label conditioning (LC). The network architecture for AMN is identical to that of the classification network for CAM with a small modification; replacing the GAP layer and the last classification layer with convolutional layers to predict pixel-level mask. Specifically, we adopt the atrous spatial pyramid pooling (ASPP) scheme~\cite{chen2017deeplab}. To generate the final pseudo-masks, we improve the predicted mask quality using the well-known refinement technique, IRN~\cite{ahn2019weakly}. Finally, we train a segmentation network with the generated pseudo-masks as supervision. Figure~\ref{fig:overview} visualizes the overall framework of AMN.

\subsection{Per-pixel classification}\label{subsection:4.2}

Based on a case study in Section~\ref{section:3}, we concluded that jointly achieving the two conditions can resolve the issue caused by the global threshold: reducing the imbalance within the foreground activation and having a large gap between the foreground and the background activations. Then, we devise an activation manipulation network (AMN) that satisfies the above two conditions. To achieve this goal, we introduce the per-pixel classification loss (PCL) because it directly enforces the two-level activation (e.g., 0 or 1), manipulating the activation (before thresholding) at the pixel-level. 

Specifically, the two-level activation as the target signal can reduce the activation imbalance inside the foreground because the foreground should be assigned to the same activation value. Likewise, the two-level activation naturally retains the large activation gap between the foreground and background. Another advantage of PCL is that it does not rely on GAP. Since GAP yields different activation maps having the same classification score, discarding the GAP can be effective in handling a global threshold problem. To train the model with per-pixel classification, we need pixel-level supervision. Under the WSSS scenario, direct access to the pixel-level supervision is prohibited. We instead utilize the refined seed $\mathbf{S}_{i}$ as noisy supervisory for training AMN. Finally, the balanced cross-entropy loss~\cite{huang2018weakly} is adopted for a per-pixel classification loss (PCL).

\subsection{Label conditioning}\label{subsection:4.3}

The original per-pixel classifier maps each pixel into one out of $N+1$ classes (i.e., a total of $N$ foreground classes and a background class). Meanwhile, label conditioning (LC) imposes that each pixel should be mapped into one out of $K+1$ classes, meaning $K$-number of classes in the ground-truth image-level label per image and $1$-background class. LC is effective in two aspects. Firstly, it helps distinguish objects with similar appearances unless they really appear together in the image. It prevents false predictions due to confusing textures (e.g., among the skin of horse, cow, or dog) by allowing activation only if its class is corresponding to any of the input ground-truth image-level label. Next, noisy pseudo-masks $\mathbf{S}_{i}$ often include a considerable number of undefined regions. The model is thus data-hungry due to lack of supervisory signals. LC can act as auxiliary supervision, providing rich learning signals. As a result, adopting LC leads to reallocating the non-target class activation to the target class activation, increasing the overall activation of the foreground. This is particularly useful to promote the less discriminative regions of the foreground.

Here, we introduce an additional layer for LC such that the effects of LC only influence high-level features. This is because limiting the choice of classes at low-level features may add to unwanted bias to the representation. Instead, we encode the ground-truth image-level labels as a feature vector and then directly multiply this vector to the feature map $f(x)$. Finally, the activation map $\mathbf{M}$ is computed as:

\vspace{-2mm}
\begin{equation}
{\small
    \mathbf{M} = \sigma(g(f(x) \cdot h(\mathbf{y}_{gt})),
}
\end{equation}\vspace{-2mm}

\noindent where $f$, $g$, and $h$ indicate a CNN backbone, convolutional layers to predict pixel-level masks, and a linear layer to map the label to the feature vector, respectively. By doing so, the feature vector of ground-truth image-level labels directly constrains the final map.

\section{Experiments}
\subsection{Experimental setup}
\noindent \textbf{Dataset \& evaluation metric.} For performance evaluation, we use both PASCAL VOC 2012~\cite{everingham2015pascal} (CC-BY 4.0) and MS COCO 2014~\cite{caesar2018coco} (CC-BY 4.0) datasets, the most popular benchmarks in the semantic segmentation task. PASCAL VOC 2012 contains 20 foreground object categories and one background category with 10,582 training images expanded by SBD~\cite{hariharan2011semantic}, 1,449 validation images, and 1,456 test images. MS COCO 2014 dataset consists of 81 classes, including a background, with 82,783 and 40,504 images for training and validation. In all experiments, we only used image-level class labels for training. For an evaluation metric, we used mean Intersection over Union (mIoU), which is widely used to measure segmentation performance.

\begin{table}
\centering 
{\small
\begin{tabular}{@{}ccccc@{}}

\toprule
CAM & PCL & LC & w/ CRF & w/ IRN~\cite{ahn2019weakly}\\
\midrule
\cmark & & & 54.3 & 66.3  \\
\cmark &\cmark & & 62.1 & 69.1  \\
\cmark &\cmark &\cmark & \textbf{65.3} & \textbf{72.2} \\
\bottomrule
\end{tabular}
}
\caption{
Ablation study of the proposed modules. The accuracy (mIoU) of pseudo-masks on PASCAL VOC 2012 train set is reported. The best score is in \textbf{bold} throughout all experiments.
}
\label{tab:ablations}
\vspace{-3mm}
\end{table}

\noindent \textbf{Implementation details.} We train the classification network to extract the seed activation map via CAM. Here, we adopt ResNet50~\cite{he2016deep} pre-trained on ImageNet~\cite{deng2009imagenet} as a backbone classification network, except for the additional layers. The CAM implementation follows the configuration from Ahn et al.~\cite{ahn2019weakly}. For training AMN, we used an Adam~\cite{kingma2014adam} optimizer and the learning rate of 5e-6 for updating the backbone parameters and 1e-4 for updating parameters associated with a per-pixel classification head. 
We adopt label smoothing as a training technique to subside the noise in initial seed, as discussed in~\cite{lukasik2020does}. The additional hyper-parameters are found in supplementary material. For the segmentation network, we experimented with DeepLab-v2 with the ResNet101 backbone~\cite{chen2017deeplab} and followed the default training settings of AdvCAM~\cite{lee2021anti}.

\subsection{Ablation study}

We investigate whether each component of AMN is effective. Considering the CAM with IRN as the baseline, we add a per-pixel classification loss (PCL) and a label conditioning (LC) in sequence and report their performances in Table~\ref{tab:ablations}. These results are from PASCAL VOC 2012 train set, thereby implying the quality of pseudo-masks. Compared to the baseline, adding PCL improves the mIoU by 2.8\%. By additionally applying LC, the performance increases by 3.1\%. Considering that the performance is already high, the additional gain by LC is impressive. Since LC suppresses any activation for non-target classes, it implicitly increases the activation of the target objects.

\begin{figure}[t]
\begin{center}
\includegraphics[]{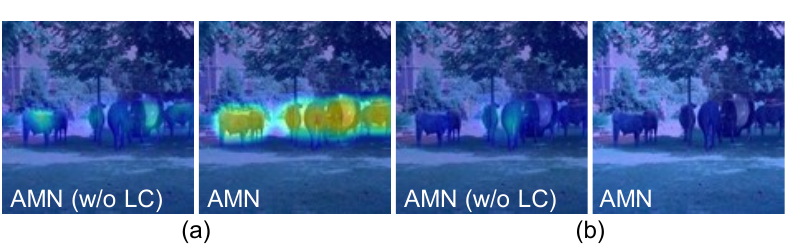}
\end{center}
\vspace{-5mm}
\caption{The effects of label conditioning on the \textit{cow} image. The wrong activations of the \textit{horse} are reallocated into the correct class when LC is applied. (a) and (b) are activation maps corresponding to \textit{cow} and \textit{horse}, respectively.}

\label{fig:label_cond}
\vspace{-5mm}
\end{figure}

To confirm the effects of LC, we visualize the activation map with and without LC in Figure~\ref{fig:label_cond}. The two classes having a similar appearance, such as \textit{cow} and \textit{horse}, can be hardly distinguishable using AMN without LC. When the image only has \textit{cow}, Figure~\ref{fig:label_cond} shows that the results without LC are activated for both the \textit{cow} and \textit{horse} (the first image of (a) and (b)). Meanwhile, after adopting LC, it is clearly seen that only \textit{cow} pixels are activated, but \textit{horse} pixels are deactivated, as shown in the second images of Figures~\ref{fig:label_cond}(a) and (b). From these results, we support that LC not only reduces non-target activations but also increases the foreground activations of the target objects. It can be interpreted that using the ground-truth image-level labels can subside the noise in our initial target (CAM with CRF), greatly increasing the performance.

We stress that LC is not applicable to the conventional image classifier because its target is already the image-level labels; LC on the image classifier can yield the model to return a trivial solution. Since AMN learns to manipulate pixel-level activation, adopting LC behaves as auxiliary supervision and leads to performance improvement.

In addition, we experimented to determine which layer of AMN is best for LC to affect. We use ResNet50 as the backbone, consisting of 4 resblocks. We applied LC after each residual block sequentially, and computed the feature vectors. The accuracy of pseudo-masks by applying LC differently is summarized in Table~\ref{tab:ablation_lc}. The best performance was achieved when LC was applied to high-level features. Likewise, the performance deteriorated when applied to low-level features. Also, applying LC on multiple layers, including the last layer, did not help the performance either. These results show that the idea of LC should be carefully implemented, because class-specific information is not always useful for feature engineering. We conjecture that the last layer handles the semantics, thus it can effectively utilize the LC for improving the final decision.

\begin{figure}[t]
\begin{center}
\includegraphics[]{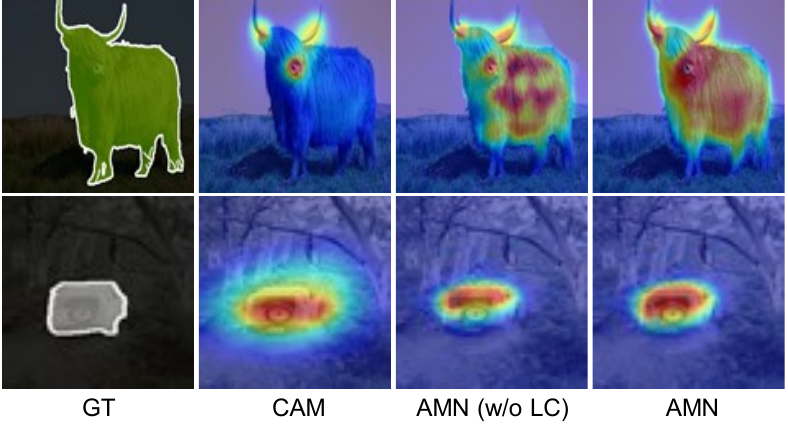}
\end{center}
\vspace{-5mm}
\caption{The effects of each component of AMN on PASCAL VOC 2012 train set. All activation maps are normalized.}
\vspace{-2mm}
\label{fig:ablation}
\end{figure}

\begin{table}
\centering 
{\small
\begin{tabular}{@{}ccccc@{}}

\toprule
layer1 & layer2 & layer3 & layer4 & mIoU\\
\midrule
\cmark &  &  &  & 48.2 \\
 & \cmark &  &  & 51.7 \\
 &  & \cmark &  & 61.2 \\
 &  &  & \cmark & \textbf{62.1} \\
 &  & \cmark & \cmark & 61.0 \\
 & \cmark & \cmark & \cmark & 51.4 \\
 & \cmark & \cmark &  & 50.9 \\

\bottomrule
\end{tabular}
}
\caption{
Accuracy (mIoU) of pseudo-masks from AMN without the boundary refinement on PASCAL VOC 2012 train set. The accuracy varies depending on where to apply LC.
}
\label{tab:ablation_lc}
\vspace{-5mm}
\end{table}

\subsection{Sensitivity to threshold}

\begin{figure}[t]
\centering
\includegraphics[width=\linewidth]{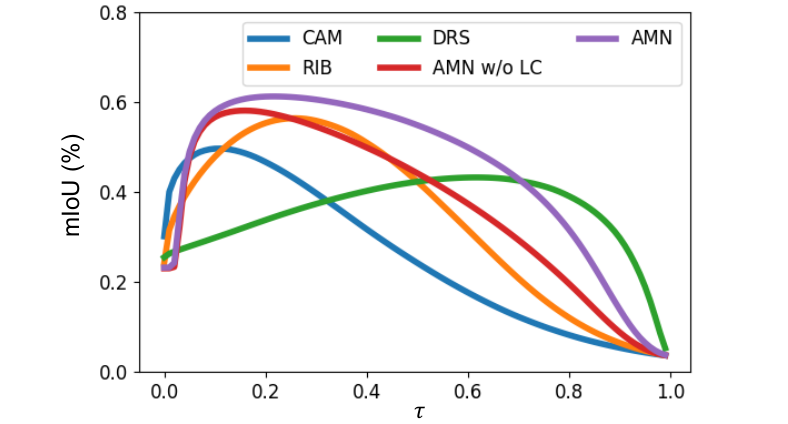}
\vspace{-6mm}
\caption{Accuracy(mIoU) of pseudo-masks depending on thresholds. The results are before boundary refinement thus differ from the final mIoU score. AMN shows more accurate and robust performance than others.}
\vspace{-3mm}
\label{fig:miou}
\end{figure}

\noindent{\textbf{Quantitative evaluation.}} In this section, we evaluate the robustness of different methods under various thresholds. We apply different thresholds to the activation map, obtain the pseudo-mask accordingly, and then measure its accuracy (mIoU). For comparison, the baseline CAM~\cite{zhou2016learning}, RIB~\cite{lee2021reducing}, DRS~\cite{KimHK21DRS} without saliency map, AMN without LC, and AMN are selected. Figure~\ref{fig:miou} presents mIoU.

In the case of CAM and RIB, the accuracy decreases even by adding a small perturbation to their global threshold; the curves fluctuate rapidly upon $\tau$. This is expected because GAP yields an imbalance in activation, thus the pseudo-mask is highly sensitive to $\tau$. On the other hand, DRS and AMN without LC exhibit relatively gentle slopes, indicating robust performances to the changes in threshold. Note that DRS tends to decrease the high activation on the most discriminative parts, thereby it partially shares the philosophy of PCL. However, DRS only focuses on suppressing foreground activations while PCL i) promotes the less discriminative parts of the foreground and ii) reduces the background activation. For this reason, activation of PCL tends to distinguish foreground and background more accurately, whereas DRS always tends to have excessive foreground coverage. These side effects of DRS finally result in relatively low accuracy in general. The original DRS utilizes a saliency map as additional supervision to compensate for this issue, thus the disadvantage was well mitigated. AMN without LC alleviated the imbalance in activation for both the foreground and background. Consequently, we achieve high mIoU and robust performance against the threshold at the same time.

Final AMN uses LC, and it increases the prior probability of a class belonging to the input image-level label. It helps reduce wrong activation for wrong classes, reallocating them to the correct class. This effectively promotes the less discriminative part of the foreground. As a result, we can achieve more accurate and robust performance. 

\begin{table}[t!]
\centering
{\small
\begin{tabular}{@{}lcc@{}}
\toprule
\multicolumn{1}{c}{Method}        &           & mIoU \\ \midrule
\multicolumn{1}{l}{IRN~\cite{ahn2019weakly}\textsubscript{CVPR'16}}     
            &  & 66.3 \\
\multicolumn{1}{l}{SEAM~\cite{wang2020self}\textsubscript{CVPR'20}}
            &  & 63.6 \\
\multicolumn{1}{l}{MBMNet~\cite{liu2020weakly}\textsubscript{ACMMM'20}}
            &  & 66.8 \\
\multicolumn{1}{l}{CONTA~\cite{zhang2020causal}\textsubscript{NeurIPS'20}}
            &  & 67.9 \\
\multicolumn{1}{l}{AdvCAM~\cite{lee2021anti}\textsubscript{CVPR'21}}
            &  & 69.9 \\ 
\multicolumn{1}{l}{RIB~\cite{lee2021reducing}\textsubscript{NeurIPS'21}}
            &  & 70.6 \\ \midrule
\multicolumn{1}{l}{AMN (ours)}
            &  & \textbf{72.2} \\ \bottomrule
\end{tabular}
}
\caption{Accuracy (mIoU) of pseudo-masks evaluated on PASCAL VOC 2012 train set.}
\label{tab:mask}
\vspace{-5mm}
\end{table}

\noindent{\textbf{Qualitative evaluation.}} Figure~\ref{fig:ablation} shows the effect of each component of AMN qualitatively. Since PCL imposes each pixel to map either the foreground or background, it penalizes the high activation in the most discriminative parts as well as the noisy activation in the background. Meanwhile, PCL can increase the moderate activation in the less discriminative part. As a natural consequence, we observe that the map generated by PCL alone (AMN without LC) covers the full extent of the object more than the original CAM. Concretely, the result from CAM concentrated on the most discriminative regions, such as a \textit{cow}'s face. Meanwhile, the resultant map by AMN without LC can capture the entire extent of the object. Depending on the object size, the CAM occasionally covers the object excessively, as seen in Figure~\ref{fig:ablation} (the bottom image for CAM). Since PCL regularizes both the foreground and background activation, it achieves reasonable coverage of object extent.  

LC provides additional supervision to undefined areas (originally less confident) and promotes the less discriminative parts of the foreground. Figure~\ref{fig:ablation} shows that activation of AMN is evenly spread inside the foreground and reasonably covers the object, such as \textit{cow} and \textit{car} region. Besides, it shows a large activation gap between the object region and the background by promoting the less discriminative regions of the foreground. As we intended, LC helps reduce activation imbalance and increases the gap between the foreground and the background activation.

\begin{figure*}[t!]
\centering
\includegraphics[width=17cm]{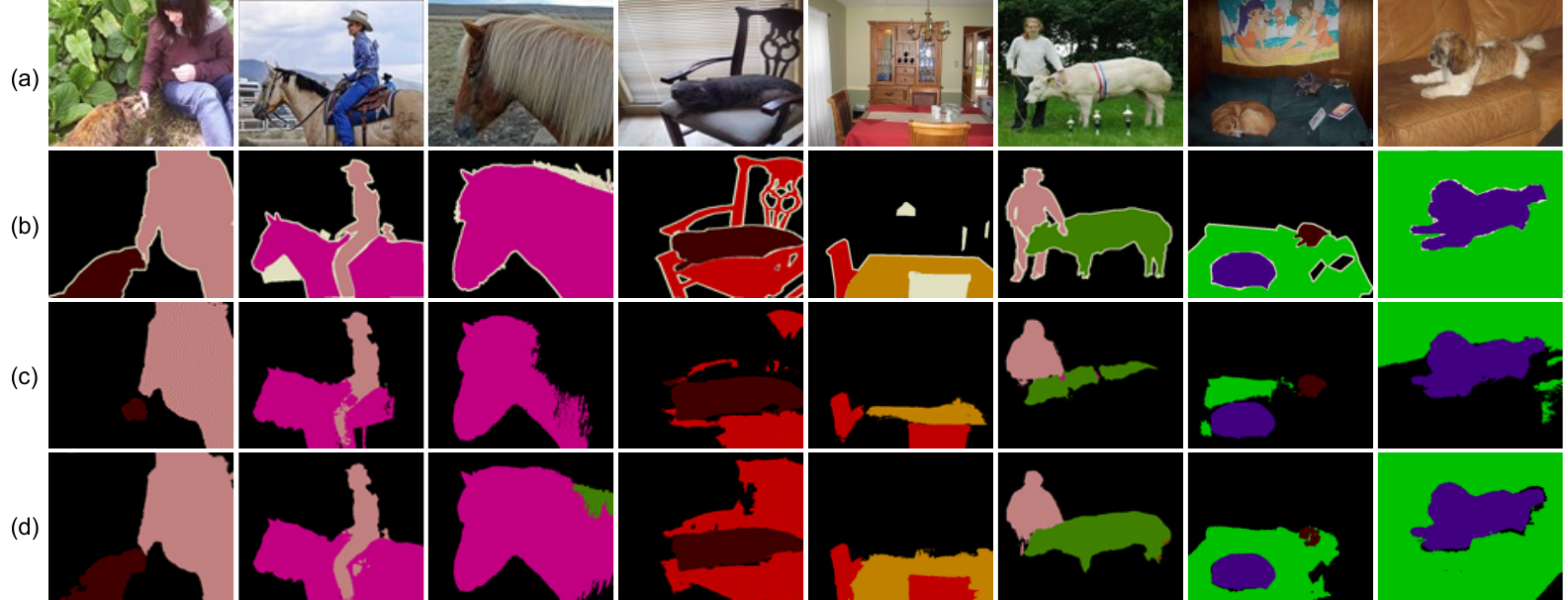}
\vspace{-2mm}
\caption{Qualitative examples of segmentation results on PASCAL VOC 2012 val set. (a) Input, (b) Ground-truth, (c) IRN, and (d) AMN.}
\label{fig:seg_qual_voc}
\vspace{-5mm}
\end{figure*}

\subsection{Comparisons with the state-of-the-arts}

\begin{table}[]
\centering
{\small
\begin{tabular}{@{}lccll@{}}
\toprule
\multicolumn{1}{c}{Method}                                                              & Backbone    & Sup.  & \multicolumn{1}{c}{val} & \multicolumn{1}{c}{test} \\ \midrule
\multicolumn{1}{l}{FickleNet~\cite{lee2019ficklenet}\textsubscript{CVPR'19}}            & ResNet101  & I.+S. & 64.9                    & 65.3                     \\
\multicolumn{1}{l}{OAA~\cite{jiang2019integral}\textsubscript{ICCV'19}}                 & ResNet101  & I.+S. & 65.2                    & 66.4                     \\
\multicolumn{1}{l}{Multi-Est.~\cite{fan2020employing}\textsubscript{ECCV'19}}           & ResNet101  & I.+S. & 67.2                    & 66.7                     \\
\multicolumn{1}{l}{MCIS~\cite{sun2020mining}\textsubscript{ECCV'20}}                    & ResNet38  & I.+S. & 66.2                    & 66.9                     \\
\multicolumn{1}{l}{SGAN~\cite{yao2020saliency}\textsubscript{ACCESS'20}}                & ResNet101  & I.+S. & 67.1                    & 67.2                     \\
\multicolumn{1}{l}{EPS~\cite{Lee_2021_CVPR}\textsubscript{CVPR'21}}                        & ResNet101  & I.+S.    & 70.9                    & 70.8                     \\ 
\multicolumn{1}{l}{DRS~\cite{KimHK21DRS}\textsubscript{AAAI'21}}                        & ResNet101  & I.+S.    & \underline{71.2}                    & \underline{71.4}                     \\ \midrule
\multicolumn{1}{l}{ICD~\cite{fan2020learning}\textsubscript{CVPR'20}}                   & ResNet101  & I.    & 64.1                    & 64.3                     \\ 
\multicolumn{1}{l}{SEAM~\cite{wang2020self}\textsubscript{CVPR'20}}                     & ResNet38  & I.    & 64.5                    & 65.7                     \\ 
\multicolumn{1}{l}{SC-CAM~\cite{chang2020weakly}\textsubscript{CVPR'20}}                & ResNet101  & I.    & 66.1                    & 65.9                     \\
\multicolumn{1}{l}{RRM~\cite{zhang2020reliability}\textsubscript{AAAI'20}}              & ResNet101  & I.    & 66.3                    & 66.5                     \\
\multicolumn{1}{l}{BES~\cite{chen2020boundary}\textsubscript{ECCV'20}}                  & ResNet101  & I.    & 65.7                    & 66.6                     \\
\multicolumn{1}{l}{CONTA~\cite{zhang2020causal}\textsubscript{NeurIPS'20}}              & ResNet50  & I.    & 66.1                    & 66.7                     \\
\multicolumn{1}{l}{AdvCAM~\cite{lee2021anti}\textsubscript{CVPR'21}}                    & ResNet101  & I.    & 68.1                    & 68.0                     \\
\multicolumn{1}{l}{RIB~\cite{lee2021reducing}\textsubscript{NeurIPS'21}}                    & ResNet101  & I.    & 68.3                    & 68.6                     \\ \midrule
\multicolumn{1}{l}{\multirow{2}{*}{AMN (ours)}}                                               & ResNet101  & I.    & 69.5$^\dagger$                    & 69.6$^\dagger$            \\
\multicolumn{1}{l}{}                                                                    & ResNet101  & I.    & \textbf{70.7}$^\ddagger$                    & \textbf{70.6}$^\ddagger$                     \\ \bottomrule
\end{tabular}
}
\vspace{-2mm}
\caption{Segmentation results (mIoU) on PASCAL VOC 2012. I. and S. denotes image-level labels and the external saliency maps used for supervision, respectively. The best score is \underline{underlined} for I.+S. and in \textbf{bold} for I. $\dagger$ for the ImageNet pre-trained model and $\ddagger$ for the MS COCO pre-trained model throughout all experiments.}
\label{tab:seg_voc}
\vspace{-5mm}
\end{table}

\noindent\textbf{Accuracy of pseudo-masks.} Similar to existing WSSS methods, we aim to improve the pseudo-mask quality and expect that it will eventually increase the accuracy of WSSS. We first evaluate the quality of pseudo-masks by comparing them with ground-truth masks. Table~\ref{tab:mask} compares mIoU of the proposed AMN with that of other state-of-the-art WSSS methods. For a fair comparison, we apply the best refinement scheme reported by each method for pseudo-mask generation. Our results achieve a gain of 5.9\% over that of IRN~\cite{ahn2019weakly}, which can be regarded as a baseline, and the gain of 1.6\% over RIB, the state-of-the-art method among the WSSS methods only with image-level labels. 

Specifically, the accuracy (mIoU) in \textit{dining table} / \textit{tv} was 41.9 / 54.2 with RIB, but 62.8 / 63.1 with ours on PASCAL VOC 2012 train set. \textit{dining table} usually exhibits extremely strong activation on the most discriminative parts (i.e., an extreme imbalance in activation), thus the optimal threshold for this class is much smaller than the global threshold. Although existing WSSS methods aim at expanding the object coverage, their effects are designed at image-level, thus cannot suppress strong activation at pixel-level. Meanwhile, AMN explicitly regularized the pixel-level activation, therefore capable of handling extreme activation. 

\begin{table}[]
\centering
{\small
\begin{tabular}{@{}lccc@{}}
\toprule
\multicolumn{1}{c}{Method}        &Backbone     &Sup.           & val \\ \midrule
\multicolumn{1}{l}{SGAN~\cite{yao2020saliency}\textsubscript{ACESS'20}}
            &VGG16      &I.+S.  & 33.6 \\
\multicolumn{1}{l}{EPS~\cite{Lee_2021_CVPR}\textsubscript{CVPR'21}}
            &VGG16      &I.+S.  & \underline{35.7} \\ \midrule
\multicolumn{1}{l}{ADL~\cite{choe2020attention}\textsubscript{TPAMI'20}}     
            &VGG16      &I.     & 30.8 \\
\multicolumn{1}{l}{CONTA~\cite{zhang2020causal}\textsubscript{NeurIPS'20}}
            &ResNet50   &I.     & 33.4 \\
\multicolumn{1}{l}{IRN~\cite{ahn2019weakly}\textsubscript{CVPR'19}}
            &ResNet101  &I.     & 41.4 \\
\multicolumn{1}{l}{RIB~\cite{lee2021reducing}\textsubscript{NeurIPS'21}}
            &ResNet101  &I.     & 43.8 \\ \midrule
\multicolumn{1}{l}{AMN (ours)}
            &ResNet101  &I.     & \textbf{44.7}$^\dagger$ \\ \bottomrule
\end{tabular}
}
\vspace{-2mm}
\caption{Accuracy (mIoU) of semantic segmentation evaluated on MS COCO 2014 val set.}
\label{tab:seg_coco}
\vspace{-5mm}
\end{table}

On the other hand, the optimal threshold for \textit{tv} significantly varies depending on the image. That means, no single threshold is meaningful. Thanks to the robust nature of AMN, we could achieve considerable gain on \textit{tv} regardless of images. These results are consistent with our motivation; threshold matters in WSSS and AMN can effectively resolve this issue. A quantitative evaluation of the pseudo-mask for each class is provided in supplementary material.

\vspace{1mm}
\noindent\textbf{Accuracy of segmentation maps.} 
For quantitative comparison, we report the mIoU scores of our method and recent WSSS methods on PASCAL VOC 2012 validation and test images. The competitors are chosen to represent the best-performing models in the last three years. On PASCAL VOC 2012 benchmark, we achieved 69.5\% and 69.6\% mIoU using the ImageNet pretrained backbone, and 70.7\% and 70.6\% mIoU with MS COCO pretrained backbone. This is a new state-of-the-art record for WSSS methods only using the image-level labels; comparable to EPS~\cite{Lee_2021_CVPR} using both image-level labels and saliency maps. 

Analogous to the observation in the pseudo-mask, our achievement is particularly affected by a large improvement in several classes that have performed poorly in the past. Specifically, the segmentation result (mIoU) in \textit{dining table} / \textit{tv} was 37.5 / 54.9 with RIB, but 53.8 / 57.5 with ours on PASCAL VOC 2012 validation set. These results are consistent with the pseudo-mask accuracy. Our method handles the strong imbalance in activation at the pixel-level (\textit{dining table}) and is robust against the threshold choice (\textit{tv}). More results on per-class mIoU scores are provided in supplementary material. Figure~\ref{fig:seg_qual_voc} shows the qualitative examples of segmentation results on PASCAL VOC 2012 validation set. These results confirm that our method covers the full extent of the objects correctly, especially \textit{dining table}, which was not possible by previous methods.

To investigate our performance on the large-scale benchmark, we adopt the MS COCO 2014 dataset. Not all competitors provide their evaluation on MS COCO 2014. For this reason, we only compare our method with five competitors. Table~\ref{tab:seg_coco} summarizes the comparison results on MS COCO 2014 validation set. AMN achieves 44.7\% mIoU, breaking a new state-of-the-art record. This demonstrates that AMN is also effective on large-scale benchmarks. More qualitative comparisons and results on per-class segmentation mIoU scores for MS COCO 2014 are in supplementary material.

\begin{figure}[t]
\begin{center}
\includegraphics[]{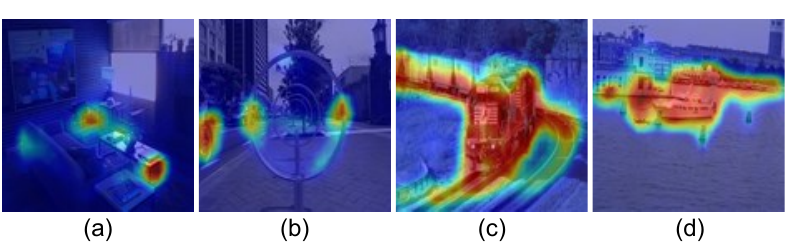}
\end{center}
\vspace{-6mm}
\caption{Failure cases of AMN. The results are normalized activation maps of (a) \textit{chair}, (b) \textit{bicycle}, (c) \textit{train}, and (d) \textit{boat}.}

\label{fig:limitation}
\vspace{-4mm}
\end{figure}

\subsection{Discussion}

\noindent \textbf{Limitation.} Although LC helps disambiguate the confusing foreground classes (e.g., visually similar to each other), it cannot handle the case where they appear together in the input image or the background is similar to the foreground object. For example, the table in Figure~\ref{fig:limitation}(a) is misclassified as chair pixels upon similar appearance. Similarly, the metal ring in Figure~\ref{fig:limitation}(b) is mispredicted as bicycle pixels due to its shared shape. In addition, our method cannot overcome the contextual bias (i.e., co-occurrence) and inaccurate boundary problem, which is inherited by CAM. Since the classifier is not designed to separate the foreground and background, activation maps from the classifier do not capture precise object boundaries, especially for complex shapes (e.g., the rough boundary of bicycles and chairs). Figures~\ref{fig:limitation}(c) and (d) show that our method cannot distinguish the co-occurring pixels in a railroad-train pair and a boat-water pair, respectively.

\noindent \textbf{Negative societal impact.} 
Since our framework consists of three training stages, it incurs more carbon emissions and power consumption. In future work, we plan to reduce the training stages, integrating the classifier and AMN.

\section{Conclusions}

In this paper, we identified that the optimal thresholds largely vary in the images, and this issue can significantly affect the performance of WSSS. To address this issue, we devised a new activation manipulation strategy for achieving robust and accurate performances. Toward this goal, we showed that jointly satisfying the two conditions can sufficiently resolve this problem. That is, we should reduce the imbalance in activation and increase the gap between the foreground and the background activation at the same time. For that, we developed an activation manipulation network (AMN) with a per-pixel classification loss and an image-level label conditioning module. Extensive experiments show that each component of AMN is effective, AMN helps induce robust pseudo-masks against the threshold, and finally achieved a new state-of-the-art performance in both PASCAL VOC 2012 and MS COCO 2014 datasets.

\noindent\textbf{Acknowledgements.} This research was supported by the NRF Korea funded by the MSIP (NRF-2022R1A2C3011154, 2020R1A4A1016619), the IITP grant funded by the MSIT (2020-0-01361/YONSEI UNIVERSITY, 2021-0-02068/Artificial Intelligence Innovation Hub), and the Korea Medical Device Development Fund grant (202011D06). 

{
    \clearpage
    \small
    \bibliographystyle{ieee_fullname}
    \bibliography{macros,main}
}

\clearpage

\appendix


\setcounter{table}{0}
\setcounter{figure}{0}
\renewcommand{\thetable}{A.\arabic{table}}
\renewcommand{\thefigure}{A.\arabic{figure}}

\twocolumn[
\centering
\Large
\textbf{Threshold Matters in WSSS: Manipulating the Activation for the Robust and Accurate Segmentation Model Against Thresholds} \\
\vspace{0.5em}Supplementary Material \\
\vspace{1.0em}
] 
\appendix
\setcounter{page}{1}
\begin{figure}[t]
\centering
\includegraphics{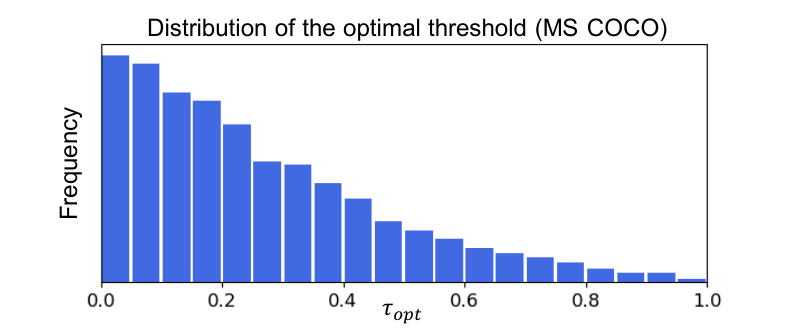}
\vspace{-1mm}
\caption{The distribution of the optimal threshold for 8,278 images randomly sampled from MS COCO 2014 train set. This shows that the optimal threshold per image quite differs from each other even on MS COCO 2014 train set.}
\label{fig:sup_per_image_th_coco}
\vspace{-5mm}
\end{figure}

\section{Implementation Details}
\noindent \textbf{Activation manipulation network.} For training AMN, we used an Adam~\cite{kingma2014adam} optimizer and the learning rate of 5e-6 for updating the backbone parameters and 1e-4 for updating parameters associated with a per-pixel classification head. Both parameter groups adopt the weight decay of 1e-4. The batch size is 16, and the total training epoch is 5. In addition, we adopted label smoothing as a training technique to subside the noise in initial seed, as discussed in~\cite{lukasik2020does}. Note that label smoothing strategy has the hyper-parameter $\epsilon$ that determines the level of smoothing (i.e., the greater $\epsilon$ indicates the stronger effect of label smoothing). In our experiment, we empirically chose $\epsilon=0.4$  and the same value was applied in all experimental settings. Specifically, given a class label $l_p \in \{0,1,2,3,...,N\}$ at pixel $p$ of the refined seed $S$, the target label distribution at $p$ is denoted as $S_p$ and it is rewritten as follows:

\begin{equation}
{\small
    S_p^c =
    \begin{cases}
    1 - \epsilon,          & c = l_p \\
    \frac{\epsilon}{N-1},  & c \neq l_p,
  \end{cases}.    
}
\end{equation}

\noindent For a per-pixel classification loss (PCL), we adopted balanced cross-entropy loss~\cite{huang2018weakly} as follows:
\vspace{0mm}
\begin{equation}
{\small
\begin{split}
    \mathcal{L}_{PCL} =
    &-{1 \over \sum_{c \in \mathcal{C}_{fg}} \vert P_{c} \vert} \sum_{c \in \mathcal{C}_{fg}} \sum_{u \in P_{c}} \log{\mathbf{M}_{u,c}} \\
    &-{1 \over \sum_{c \in \mathcal{C}_{bg}} \vert P_{c} \vert} \sum_{c \in \mathcal{C}_{bg}} \sum_{u \in P_{c}} \log{\mathbf{M}_{u,c}},    
\end{split}}
\end{equation}
\begin{equation}
{\small
\begin{split}
    \mathbf{M} = \sigma(g(f(x))),
\end{split}} \label{eq:mask}
\end{equation}\vspace{-2mm}

\noindent where $\sigma$ is the softmax function, $\mathbf{M}$ is the activation map from AMN ($\mathrm{F}_c$ is the class activation map from the classifier), $\mathcal{C}_{fg}$ is the set of classes that are present in the image (excluding background) and $\mathcal{C}_{bg}$ is the background class. $|P_{c}|$ denotes the number of pixels belonging to class $c$.

\vspace{1mm}

\noindent \textbf{Segmentation network.} 
For the segmentation network, we adopted DeepLab-v2-ResNet101 and followed the default training settings of AdvCAM~\cite{lee2021anti} for PASCAL VOC 2012. Input images are randomly scaled to $[0.5, 0.75, 1.0, 1.25, 1.5, 1.75, 2.0]$ and cropped to $321 \times 321$ ($481 \times 481$ for MS COCO 2014) for training. We used the SGD optimizer with the batch size of 10 (20 for MS COCO 2014), the momentum of $0.9$, and the weight decay of $5e-4$. The number of training iterations is 30k and the initial learning rate is $2.5e-4$ with the polynomial learning rate decay $lr_{iter} = lr_{init}(1 - \frac{iter}{max_iter})^\gamma$, where $\gamma$ is set to $0.9$. We used balanced cross-entropy loss~\cite{huang2018weakly} as in AdvCAM~\cite{lee2021anti}.

\section{Additional Analysis}
\noindent \textbf{Distribution of optimal thresholds on MS COCO 2014.} Figure~\textcolor{red}{1} shows the distribution of optimal threshold in the PASCAL VOC 2012. Here, we further investigate whether the same observation holds in MS COCO 2014, which is a large-scale, popular benchmark dataset for semantic segmentation. To efficiently derive the distribution of optimal threshold using MS COCO 2014, we randomly sample 10\% of MS COCO 2014 and find the optimal threshold for each image. Figure~\ref{fig:sup_per_image_th_coco} shows that the optimal threshold per image is distributed over a wide range from 0 to 1. This result confirms that our observation in PASCAL VOC 2012 consistently holds in a different dataset; the global threshold is not sufficient to generate the optimal pseudo-masks. 

\vspace{1mm}

\noindent \textbf{Effects of encoding features.} In Section~\textcolor{red}{4.3}, we encode label vectors by transforming it into feature vectors for label conditioning. To differentiate the effect of label vector from the effect of encoding any vectors, we conduct additional experiments; 1) encoding a one-vector, 2) encoding the label vector + a random vector and 3) encoding the label vector. Table~\ref{tab:lc_embedded} compares three cases by reporting the accuracy (mIoU) of pseudo-masks. With a one-vector, no distinct gain is observed over AMN without LC. This implies that the encoding operation itself does not make much difference. In addition, we observe the accuracy gain when encoding noisy labels (i.e., the ground-truth label vector summed up with a noise vector). Since this noisy label also reduces the possible choices, it helps reduce non-target activation to some extent. As expected, the ground-truth image-level labels can lead a noticeable gain, achieving the best accuracy among all.  

\begin{table}
\centering 
{\small
\begin{tabular}{@{}ccccc@{}}

\toprule
 & AMN    & AMN & AMN & \multirow{2}{*}{AMN}\\
 & w/o LC & w/ ones & w/ label + noise & \\
\midrule
mIoU & 58.2\% & 58.6\% & 60.5\% & 62.1\% \\
\bottomrule
\end{tabular}
}
\caption{
Accuracy (mIoU) of pseudo-masks from AMN without the boundary refinement on PASCAL VOC 2012 train set. The accuracy depends on the information encoded through the label conditioning module.
}
\label{tab:lc_embedded}
\end{table}

\begin{figure}[t]
\centering
\includegraphics{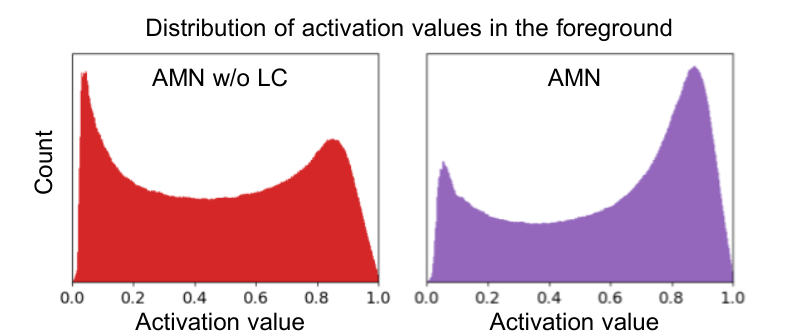}
\vspace{-1mm}
\caption{The distribution of activation values in the foreground on PASCAL VOC 2012 train set. This shows that LC not only reduces non-target activations but also increase the foreground activations of the target objects.}
\label{fig:sup_fg_activations}
\vspace{-5mm}
\end{figure}

\noindent \textbf{Effect of label conditioning.} Additionally, we observe the histogram of foreground activation values on PASCAL VOC 2012 train set. For this empirical study, we focus on the activation values appearing inside the target objects using ground-truth segmentation mask. As shown in Figure~\ref{fig:sup_fg_activations}, the effects of LC increase the foreground activations of the target objects--the values within [0.8 1.0] greatly increase and the values within [0.0 0.2] sufficiently decrease. This is coherent with our observation in Figure~\textcolor{red}{4}, where LC reduces the horse activation in the cow image and then the cow is correctly activated after applying LC. Overall, we confirm that LC is effective to achieve accurate and robust segmentation performance.

\section{Per-class Performance}
\begin{figure*}[t]
\centering
\includegraphics{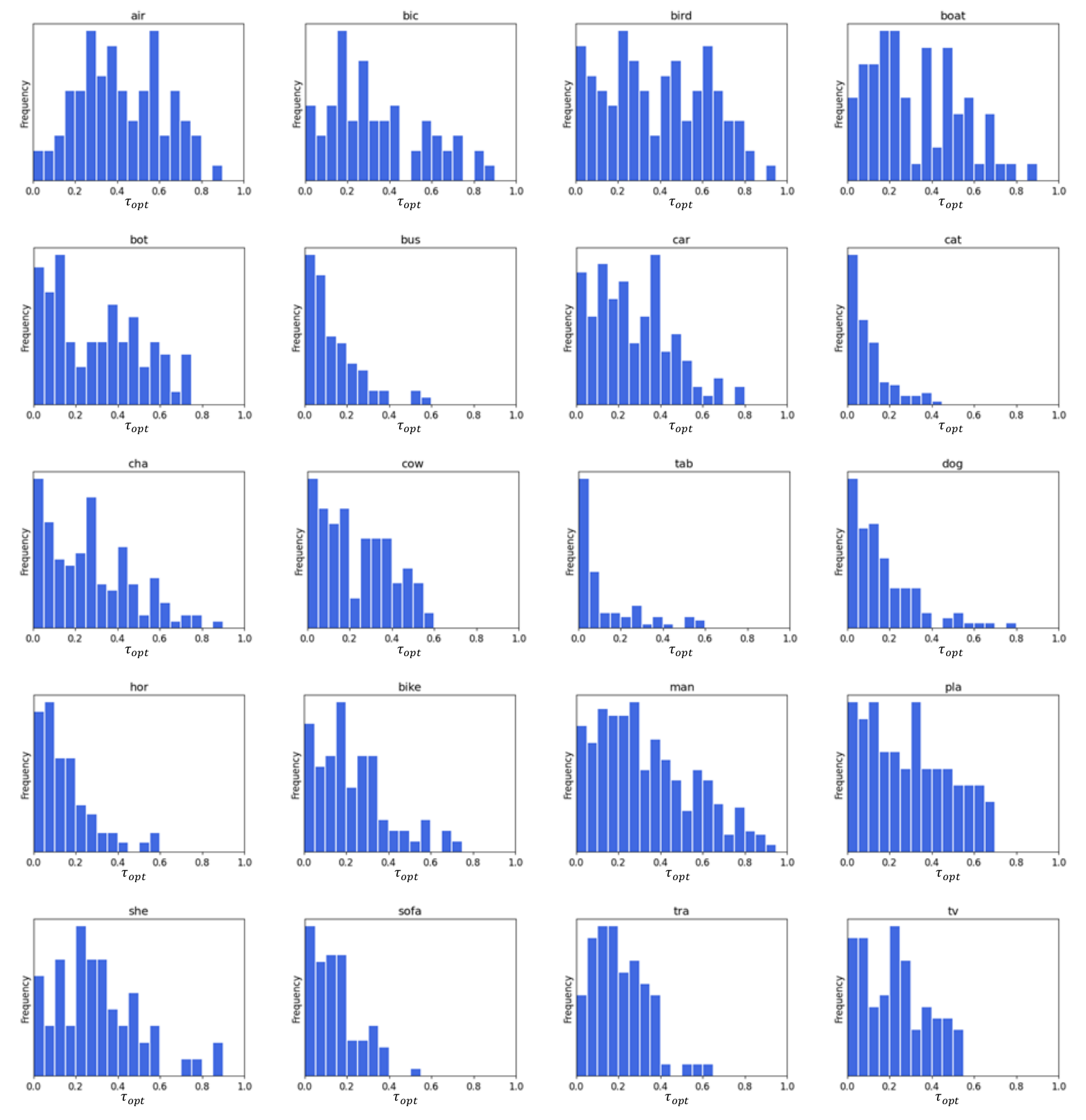}
\vspace{-1mm}
\caption{The distribution of the optimal threshold per class on PASCAL VOC 2012 train set. This shows that the distribution of the optimal threshold per class is quite different.}
\label{fig:sup_per_class_th_voc}
\vspace{-5mm}
\end{figure*}

In Figure~\textcolor{red}{1}(a), we showed that the optimal threshold per image quite differs from each other. Herein, Figure~\ref{fig:sup_per_class_th_voc} shows the distribution of the optimal threshold per image within the same class on PASCAL VOC 2012 train set. From these results, we find that the distribution of the optimal threshold is widely distributed in most classes and the different class has different tendency; a class-wise global threshold is also largely different from each other. 

\begin{figure*}[t]
\centering
\includegraphics{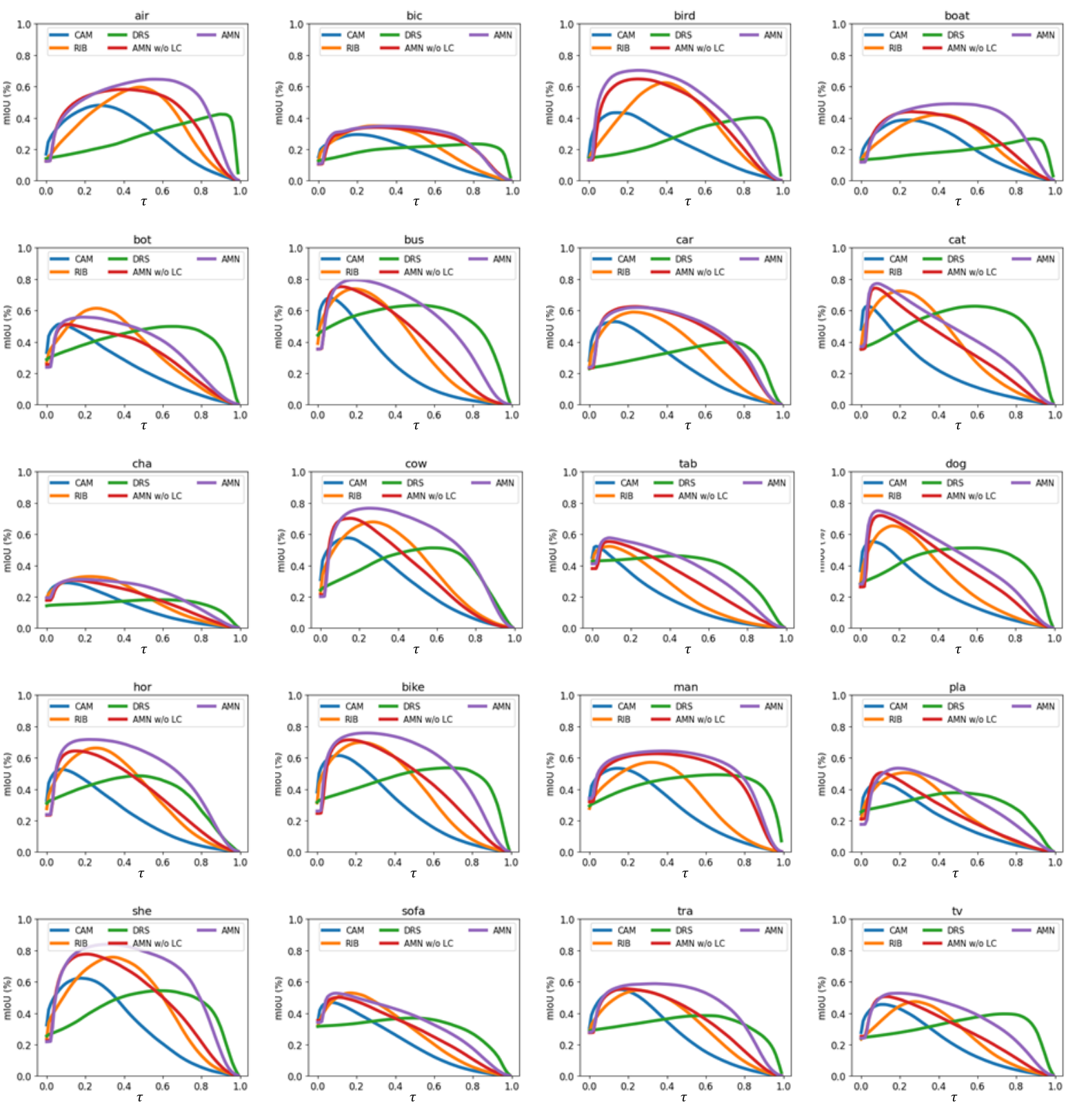}
\vspace{-1mm}
\caption{Per-class mIoU of pseudo-masks according to thresholds on PASCAL VOC 2012 train set. The results are before boundary refinement. AMN shows generally more accurate and robust performance than others.}
\label{fig:sup_per_class_miou_voc}
\vspace{-5mm}
\end{figure*}

Figure~\ref{fig:sup_per_class_miou_voc} shows per-class mIoU of the pseudo-masks according to thresholds on PASCAL VOC 2012 train set. Although the different class exhibits different characteristics in optimal thresholds, AMN tends to generate more accurate and robust pseudo-masks (e.g., the pseudo-mask accuracy of \textit{man} increases a lot, but that of \textit{sofa} is almost same).

Table~\ref{per_class_pseudo_mask_voc} shows the per-class mIoU of the pseudo-mask results on PASCAL VOC 2012 train set. For comparison, we report the per-class mIoU of RIB~\cite{lee2021reducing}. Since RIB does not present the per-class mIoU of the pseudo-masks, we reproduced their results based on the official implementation of RIB\footnote{\url{https://github.com/jbeomlee93/RIB}}. Table~\ref{per_class_pseudo_mask_voc} and Table~\ref{per_class_seg_coco} show the per-class mIoU of the segmentation results on PASCAL VOC 2012 and MS COCO 2014 datasets, respectively. Specifically, for MS COCO 2014 validation set, we observe the strong gains in several classes; \textit{dining table} / \textit{airplane} are 11.6 / 61.3 with RIB, but 17.2 / 65.5 with ours. These results are consistent with the PASCAL VOC 2012; our method handles the strong imbalance in activation at the pixel-level (\textit{dining table}) and is robust against the threshold choice (\textit{airplane}). This demonstrates that AMN is also effective on large-scale benchmarks.

\section{Qualitative Examples}
\begin{figure*}[t!]
\centering
\includegraphics[width=17cm]{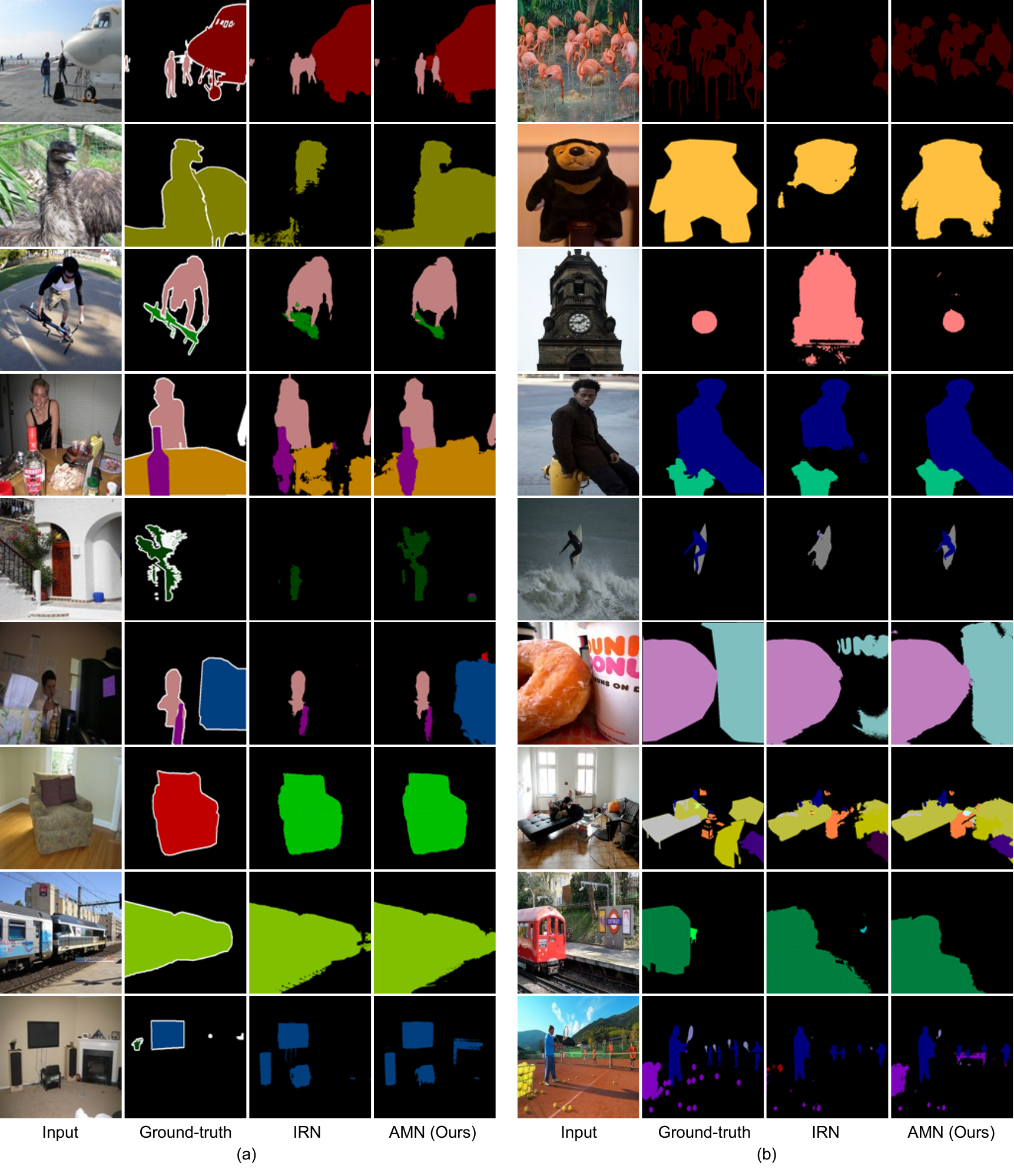}
\vspace{-2mm}
\caption{Qualitative examples of segmentation results on (a) PASCAL VOC 2012 val set and (b) MS COCO 2014 val set.}
\label{fig:sup_qual_seg}
\end{figure*}

Figure~\ref{fig:sup_qual_seg} shows qualitative examples and failure cases of segmentation results from AMN on PASCAL VOC 2012 validation set and MS COCO 2014 validation set. Our method effectively covers the full extent of the objects. Meanwhile, we still have some failure cases: 1) confusing objects (e.g., \textit{sofa} and \textit{chair}), 2) co-occurrence problem (e.g., \textit{railroad} and \textit{train}, 3) shape bias (e.g., \textit{tv/monitor}).


\renewcommand{\tabcolsep}{2pt}

\begin{table*}[t]
  \centering
  \begin{adjustbox}{max width=\textwidth}
    \begin{tabular}{lcccccccccccccccccccccc}
    
        \Xhline{1pt}

        \\[-0.95em]
     & bkg& aero  & bike  & bird  & boat  & bottle & bus   & car   & cat   & chair & cow   & table & dog   & horse & motor & person & plant & sheep & sofa  & train & tv  ~ & mIOU \\
   
    \midrule
    RIB$^*$     &88.9	&70.3	&44.5	&74.5	&62.3	&77.8	&83.3	&73.9	&85.9	&40.8	&82.4	&41.9	&79.7	&83.4	&80.6	&69.0	&59.5	&83.7	&63.9	&60.8	&54.2	&69.6 \\
    AMN (Ours) &90.2	&75.3	&40.1	&77.4	&67.9	&73.4	&85.6	&78.9	&80.7	&36.5	&86.1	&62.8	&78.7	&83.4	&81.0	&74.4	&62.4	&89.4	&62.8	&65.3	&63.1	&72.2 \\
    \bottomrule
    \vspace{-2em}
    \end{tabular}%
  \end{adjustbox}%
  \caption{Per-class accuracy (mIoU) of pseudo-masks evaluated on PASCAL VOC 2012 train set. $^*$ denotes the reproduced results based on the official implementation of RIB~\cite{lee2021reducing}.}
  \vspace{-1em}
  
  \label{per_class_pseudo_mask_voc}%
\end{table*}%
\renewcommand{\tabcolsep}{2pt}

\begin{table*}[t]
  \centering
  \begin{adjustbox}{max width=\textwidth}
    \begin{tabular}{lcccccccccccccccccccccc}
    
        \Xhline{1pt}

        \\[-0.95em]
     & bkg& aero  & bike  & bird  & boat  & bottle & bus   & car   & cat   & chair & cow   & table & dog   & horse & motor & person & plant & sheep & sofa  & train & tv  ~ & mIOU \\
   
    \midrule
    \multicolumn{22}{l}{Results on PASCAL VOC 2012 val set:}\\
    AdvCAM  &90.0	&79.8	&34.1	&82.6	&63.3	&70.5	&89.4	&76.0	&87.3	&31.4	&81.3	&33.1	&82.5	&80.8	&74.0	&72.9	&50.3	&82.3	&42.2	&74.1	&52.9	&68.1 \\
    RIB     &90.3	&76.2	&33.7	&82.5	&64.9	&73.1	&88.4	&78.6	&88.7	&32.3	&80.1	&37.5	&83.6	&79.7	&75.8	&71.8	&47.5	&84.3	&44.6	&65.9	&54.9	&68.3 \\
    AMN (Ours) &90.6	&79.0	&33.5	&83.5	&60.5	&74.9	&90.0	&81.3	&86.6	&30.6	&80.9	&53.8	&80.2	&79.6	&74.6	&75.5	&54.7	&83.5	&46.1	&63.1	&57.5	&69.5 \\
    \midrule
    \multicolumn{22}{l}{Results on PASCAL VOC 2012 test set:}\\
    AdvCAM  &90.1	&81.2	&33.6	&80.4	&52.4	&66.6	&87.1	&80.5	&87.2	&28.9	&80.1	&38.5	&84.0	&83.0	&79.5	&71.9	&47.5	&80.8	&59.1	&65.4	&49.7	&68.0 \\
    RIB     &90.4	&80.5	&32.8	&84.9	&59.4	&69.3	&87.2	&83.5	&88.3	&31.1	&80.4	&44.0	&84.4	&82.3	&80.9	&70.7	&43.5	&84.9	&55.9	&59.0	&47.3	&68.6 \\    
    AMN (Ours) &90.7	&82.8	&32.4	&84.8	&59.4	&70.0	&86.7	&83.0	&86.9	&30.1	&79.2	&56.6	&83.0	&81.9	&78.3	&72.7	&52.9	&81.4	&59.8	&53.1	&56.4	&69.6 \\
    \bottomrule
    \vspace{-2em}
    \end{tabular}%
  \end{adjustbox}%
  \caption{Per-class accuracy (mIoU) of segmentation results evaluated on PASCAL VOC 2012.}
  \label{per_class_seg_coco}%
\end{table*}%
\begin{table*}[htbp]
  \centering
  \normalsize
  \begin{adjustbox}{max width=\textwidth}
    \begin{tabular}{lc@{\hskip 0.1in}c@{\hskip 0.1in}c@{\hskip 0.1in}|@{\hskip 0.1in}lc@{\hskip 0.1in}c@{\hskip 0.1in}c@{\hskip 0.1in}|@{\hskip 0.1in}lc@{\hskip 0.1in}c@{\hskip 0.1in}c@{\hskip 0.1in}|@{\hskip 0.1in}lc@{\hskip 0.1in}c@{\hskip 0.1in}c@{\hskip 0.1in}|@{\hskip 0.1in}lc@{\hskip 0.1in}c@{\hskip 0.1in}c}
    \Xhline{1pt}
    Class & IRN & RIB & Ours & Class & IRN & RIB & Ours & Class & IRN & RIB & Ours & Class & IRN & RIB & Ours & Class & IRN & RIB & Ours \\
    \hline
    \hline
    
    background      & 80.5 & 82.0 & 82.8 &  dog          & 56.2 & 63.5 & 67.9 &  kite            & 28.8 & 37.1 & 43.9 &  broccoli       & 52.6 & 45.4 & 45.9 &  cell phone      & 51.6 & 54.1 & 57.7 \\
    person          & 45.9 & 56.1 & 53.7 &  horse        & 58.1 & 63.6 & 65.3 &  baseball bat    & 12.6 & 15.3 & 16.1 &  carrot         & 37.0 & 34.6 & 31.3 &  microwave       & 42.7 & 45.2 & 43.2 \\
    bicycle         & 48.9 & 52.1 & 49.3 &  sheep        & 64.6 & 69.1 & 71.9 &  baseball glove  & 7.9  & 8.1  & 6.5  &  hot dog        & 48.4 & 49.7 & 47.0 &  oven            & 31.0 & 35.9 & 35.5 \\
    car             & 31.3 & 43.6 & 38.9 &  cow          & 63.8 & 68.3 & 70.3 &  skateboard      & 27.1 & 31.8 & 29.6 &  pizza          & 55.9 & 58.9 & 57.5 &  toaster         & 16.4 & 17.8 & 24.3 \\
    motorcycle      & 64.7 & 67.6 & 67.1 &  elephant     & 79.3 & 79.5 & 81.4 &  surfboard       & 40.7 & 29.2 & 44.6 &  donut          & 50.0 & 53.1 & 57.3 &  sink            & 33.3 & 33.0 & 31.4 \\
    airplane        & 62.0 & 61.3 & 65.5 &  bear         & 74.6 & 76.7 & 79.9 &  tennis racket   & 49.7 & 48.9 & 45.6 &  cake           & 38.6 & 40.7 & 40.1 &  refrigerator    & 40.0 & 46.0 & 45.6 \\
    bus             & 60.4 & 68.5 & 68.1 &  zebra        & 79.7 & 80.2 & 82.4 &  bottle          & 30.9 & 33.1 & 33.0 &  chair          & 17.7 & 20.6 & 23.6 &  book            & 29.9 & 31.1 & 29.5 \\
    train           & 51.1 & 51.3 & 56.3 &  giraffe      & 72.3 & 74.1 & 76.5 &  wine glass      & 24.3 & 27.5 & 31.7 &  couch          & 32.6 & 36.8 & 36.6 &  clock           & 41.3 & 41.9 & 47.6 \\
    truck           & 32.2 & 38.1 & 38.9 &  backpack     & 19.1 & 18.1 & 15.5 &  cup             & 27.3 & 27.4 & 28.8 &  potted plant   & 10.5 & 17.0 & 19.2 &  vase            & 28.4 & 27.5 & 30.9 \\
    boat            & 36.7 & 42.3 & 41.6 &  umbrella     & 57.3 & 60.1 & 62.4 &  fork            & 16.9 & 15.9 & 16.3 &  bed            & 33.8 & 46.2 & 44.5 &  scissors        & 41.2 & 41.0 & 39.2 \\
    traffic light   & 48.7 & 47.8 & 49.6 &  handbag      & 9.0  & 8.6  & 7.2  &  knife           & 15.6 & 14.3 & 16.3 &  dining table   & 6.7  & 11.6 & 17.2 &  teddy bear      & 56.4 & 62.0 & 63.9 \\
    fire hydrant    & 74.9 & 73.4 & 74.3 &  tie          & 24.0 & 28.6 & 28.7 &  spoon           & 8.4  & 8.2  & 8.4  &  toilet         & 63.4 & 63.9 & 65.4 &  hair drier      & 16.2 & 16.7 & 21.3 \\
    stop sign       & 76.8 & 76.3 & 70.8 &  suitcase     & 45.2 & 49.2 & 48.6 &  bowl            & 17.0 & 20.7 & 24.4 &  tv             & 35.5 & 39.7 & 43.5 &  toothbrush      & 16.7 & 21.0 & 25.0 \\
    parking meter   & 67.3 & 68.3 & 63.2 &  frisbee      & 53.8 & 53.6 & 56.6 &  banana          & 62.4 & 59.8 & 61.1 &  laptop         & 39.3 & 48.2 & 51.8 &  \multicolumn{1}{c}{} & \multicolumn{1}{c}{} & \multicolumn{1}{c}{}\\
    \cline{17-20}
    bench           & 31.4 & 39.7 & 35.0 &  skis         & 8.0  & 9.7  & 11.4 & apple            & 43.3 & 48.5 & 45.9 &  mouse          & 27.9 & 22.4 & 30.0 &  \multicolumn{1}{l}{\multirow{4}[0]{*}{mean}} & \multicolumn{1}{l}{\multirow{4}[0]{*}{41.4}} & \multicolumn{1}{l}{\multirow{4}[0]{*}{43.8}} & \multicolumn{1}{l}{\multirow{4}[0]{*}{44.7}}\\
    bird            & 55.5 & 57.5 & 60.0 &  snowboard    & 25.5 & 29.4 & 30.3 & sandwich         & 37.9 & 36.9 & 35.8 &  remote         & 41.4 & 38.0 & 38.4 &  \multicolumn{1}{c}{} & \multicolumn{1}{c}{} & \multicolumn{1}{c}{} \\
    cat             & 68.2 & 72.4 & 71.2 &  sports ball  & 33.6 & 38.0 & 33.9 &  orange          & 60.1 & 62.5 & 62.9 &  keyboard       & 52.9 & 50.9 & 48.7 &  \multicolumn{1}{c}{} & \multicolumn{1}{c}{} & \multicolumn{1}{c}{} \\
    \Xhline{1pt}
    \vspace{-2em}
    \end{tabular}%
    \end{adjustbox}%
  \caption{Per-class accuracy (mIoU) of segmentation results evaluated on MS COCO 2014.}
  \label{per_class_seg_coco}%
\end{table*}%

\end{document}